\documentclass{article}

\pdfoutput=1

\usepackage[preprint,nonatbib]{nips_2018}

\usepackage[utf8]{inputenc} 
\usepackage[T1]{fontenc}    
\usepackage{hyperref}       
\usepackage{url}            
\usepackage{booktabs}       
\usepackage{amsmath}
\usepackage{amsfonts}       
\usepackage{nicefrac}       
\usepackage{microtype}      
\usepackage{graphicx}
\usepackage{subcaption}
\usepackage{todonotes}
\usepackage{caption}
\usepackage{multirow}
\usepackage{diagbox}

\usepackage[symbol]{footmisc}

\newcommand{\bx}{\mathbf{x}}
\newcommand{\Sh}{\hat{S}}
\newcommand{\Nfg}{N_\text{fg}}
\newcommand{\Nbg}{N_\text{bg}}
\newcommand{\LL}{\mathcal{L}}
\newcommand{\LLfg}{\mathcal{L}_\text{fg}}
\newcommand{\LLbg}{\mathcal{L}_\text{bg}}

\definecolor{topiccolor}{rgb}{0, 0.5, 0}

\title{One-shot Texture Segmentation}
\date{\today}

\author{
  Ivan Ustyuzhaninov\\
  University of T\"ubingen\\
  \texttt{ivan.ustyuzhaninov@bethgelab.org} \\
  \And
  Claudio Michaelis\\
  University of T\"ubingen\\
  \texttt{claudio.michaelis@bethgelab.org} \\
  \And
  Wieland Brendel\thanks{co-senior authors}\\
  University of T\"ubingen\\
  \texttt{wieland.brendel@bethgelab.org} \\
  \And
  Matthias Bethge\footnotemark[1] \\
  University of T\"ubingen \\
  \texttt{matthias@bethgelab.org} \\
}

\begin{document}

\maketitle

\begin{abstract}
    We introduce one-shot texture segmentation: the task of segmenting an input image containing multiple textures given a patch of a reference texture. This task is designed to turn the problem of texture-based perceptual grouping into an objective benchmark. We show that it is straight-forward to generate large synthetic data sets for this task from a relatively small number of natural textures. In particular, this task can be cast as a self-supervised problem thereby alleviating the need for massive amounts of manually annotated data necessary for traditional segmentation tasks. In this paper we introduce and study two concrete data sets: a dense collage of textures (CollTex) and a cluttered texturized Omniglot data set. We show that a baseline model trained on these synthesized data is able to generalize to natural images and videos without further fine-tuning, suggesting that the learned image representations are useful for higher-level vision tasks.
\end{abstract}

\begin{figure}[h]
    \centering
    \begin{subfigure}{0.48\textwidth}
        \includegraphics[width=\linewidth]{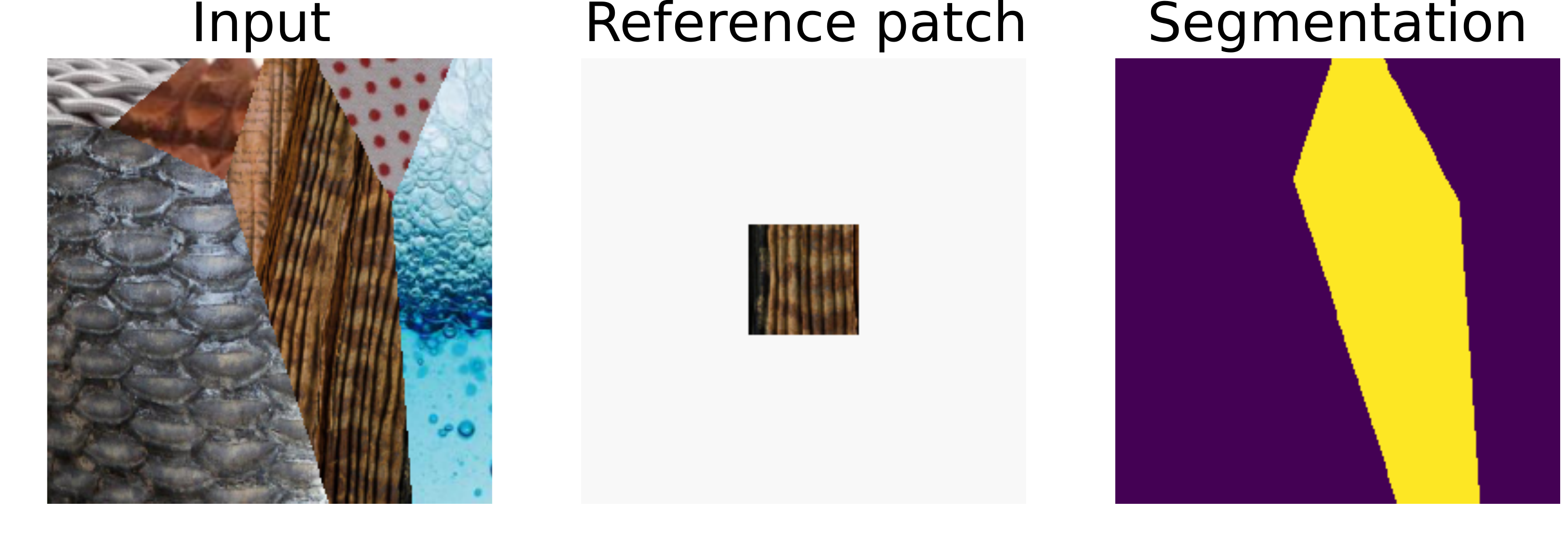}
        \caption{Collage of natural textures}
    \end{subfigure}
    \hfill
    \begin{subfigure}{0.48\textwidth}
        \includegraphics[width=\linewidth]{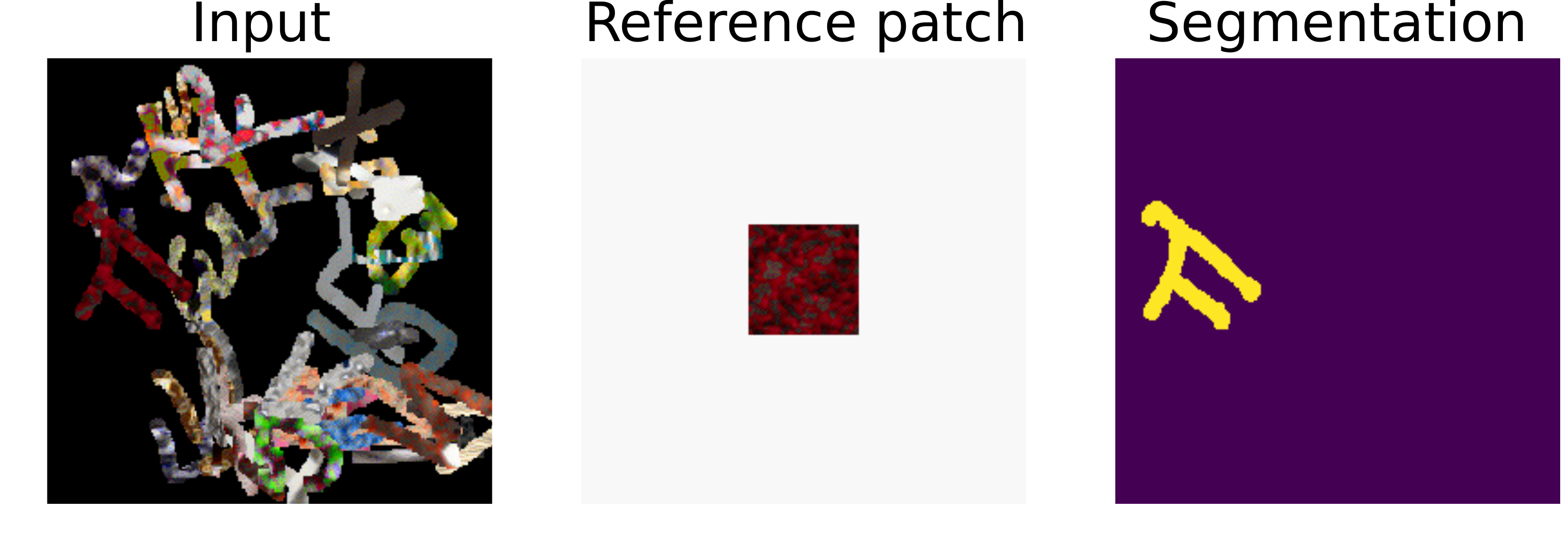}
        \caption{32 Omniglot characters with overlayed textures}
    \end{subfigure}
    \caption{Examples of texture segmentation tasks. In each example the three images correspond to input image, reference texture and target segmentation (from left to right).}
    \label{fig:figure1}
\end{figure}

\section{Introduction}

The human visual system is remarkably robust to many local image transformations. There is evidence that outside of the foveal region the visual system represents image patches as textures \cite{wallis2017, freeman2011} which are robust to local shifts and deformations. Such a compression can be extremely useful as it summarizes the mostly irrelevant fine-scale details (e.g. the individual blades of grass) but keeps the rough low-level semantic concepts (e.g. grass). This suggests that local texture features can form an important mid-level image representation.

To learn such representations we suggest the task of one-shot texture segmentation: given an image containing multiple textures as well as a reference texture, the task is to segment parts of the input image covered by the reference texture (see Figure~\ref{fig:figure1}). Since these parts can be highly variable in size and shape, models have to learn an extremely flexible and robust texture representation that can finely discriminate the local image statistics. Such a problem formulation allows for algorithmic data set generation which in turn allows to turn the underlying problem of perceptual texture grouping into a supervised learning task.

Our contributions are as follows:
\begin{itemize}
    \item We introduce and openly release\footnote[2]{\url{https://github.com/ivust/one-shot-texture-segmentation}} two benchmark data sets for one-shot texture segmentation as well as the code necessary to generate the data in a self-supervised fashion (Section~\ref{sec:one-shot-segmentation-tasks}).
    \item We introduce and train a strong baseline segmentation model for this task (Sections~\ref{sec:proposed-architecture}, \ref{sec:loss-training-evaluation} and \ref{subsec:results-best-models}).
    \item We show that one-shot texture segmentation requires computations of local image statistics of higher-orders. (Section~\ref{subsec:non-trivial-embedding}).
    \item We demonstrate that the texture representations learned by our model generalize to natural images and can be used to obtain coarse semantic segmentations of natural images and videos without additional fine-tuning (Sections~\ref{subsec:results-video-segmentation} and \ref{subsec:results-object-classification}).
\end{itemize}

\section{Related work}
\label{sec:related-work}

\paragraph{Segmentation} 
Segmentation tasks can be roughly split into Semantic Segmentation \cite{long2015, chen2015, ronneberger2015} (including the recently introduced class-agnostic segmentation \cite{maninis2018, papadopoulos2017}) and Instance Segmentation \cite{pinheiro2015, mask-r-cnn}. Popular models for this task are encoder-decoder architectures with skip connections \cite{badrinarayanan2017, ronneberger2015} which are often based on pretrained DNNs like VGG \cite{vgg} as feature extractors. Virtually all existing segmentation tasks \cite{pascal_voc, mscoco, zhou2017} are based on manual human labels which are usually hard to get \cite{zhou2017, zhu2017, acuna2018, barruiso2012} and concentrate on high-level semantic concepts. In contrast, we here focus on segmentations defined by mid-level texture cues for which tasks can be defined in a self-supervised way.

\paragraph{One-Shot Learning} While the corpus of work on one-shot learning is quite extensive \cite{omniglot2015, koch2015, vinyals2016, snell2017, triantafillou2017, qiao2018, cai2018, gidaris2018}, more complex tasks like one-shot segmentation \cite{shaban2017, rakelly2018} have been established only recently. Our work is most similar to the Omniglot one-shot segmentation setting recently introduced by \cite{claudio2018} in which cluttered Omniglot images are segmented given a reference shape. However, instead of segmenting based on shape similarity, we here focus on segmentations defined by the texture of the reference patch.

\paragraph{Texture modelling} Computational modelling of textures has been a long-standing task in computer vision. Parametric models based on image statistics ($N$-th order joint histograms) were introduced by \cite{julesz1962}, followed by models based on other statistical measurements \cite{heeger1995, portilla2000}. Current state-of-the-art parametric texture models are based on DNN features \cite{gatys2015, liu2016}. We build on these works and use DNNs as extractors of texture parameters to build our model. Another line of work in texture modelling focuses on non-parametric methods \cite{efros1999, levina2006}, which are less suited for our approach since we rely on parametric representations to match textures in input and reference images. 

\paragraph{Texture Perception} Textures are well-suited for studying human perception as they are diverse, non-trivial stimuli. There is evidence that certain areas of the human brain are building texture representations to perform low-level visual inference \cite{yu2015, motoyoshi2007}. Within the substantial body of work on texture modelling \cite{victor2017}, experimental settings very similar to ours have been introduced. For instance, in \cite{victor2015, harrison2008} human subjects have to perform texture segmentation of images consisting of multiple textures, relying on local image statistics in a cluttered scene. In \cite{serre2007} a biologically motivated model of texture perception has been used to recognize objects. Finally, textures have also been used to study the difference between human perception and computational models of vision \cite{grossberg1985}.

\section{One-shot texture segmentation data sets}
\label{sec:one-shot-segmentation-tasks}

In the following section we describe two data sets for one-shot texture segmentation that can be algorithmically generated from a set of unlabelled natural texture images. We use textures from the Describable Textures Data set (DTD, 5640 textures) \cite{dtd} and synthesized ones from a VGG texture model \cite{gatys2015}. The input images are of the size $256 \times 256$ and contain multiple textures. One of these textures is presented as a $64 \times 64$ pixels reference image. The task for all data sets is to segment out the parts of the input image covered by the reference texture.

\subsection{Collages of natural textures (CollTex)}
\label{subsec:colltex}

Our first data set consists of collages of natural textures. Each input image is partitioned randomly into non-overlapping regions which are filled with randomly selected textures from a fixed training set of texture images (DTD, see examples in Figures~\ref{fig:figure1} and \ref{fig:results}). To generate random partitions, we uniformly sample $N$ anchor points at random positions in the image, and then assign each image pixel to one of the $N$ segmentation regions defined by the closest anchor point. We generate samples in real-time and reserve 100 randomly chosen DTD textures for testing.

\begin{figure}[t]
    \centering
    \includegraphics[width=\textwidth]{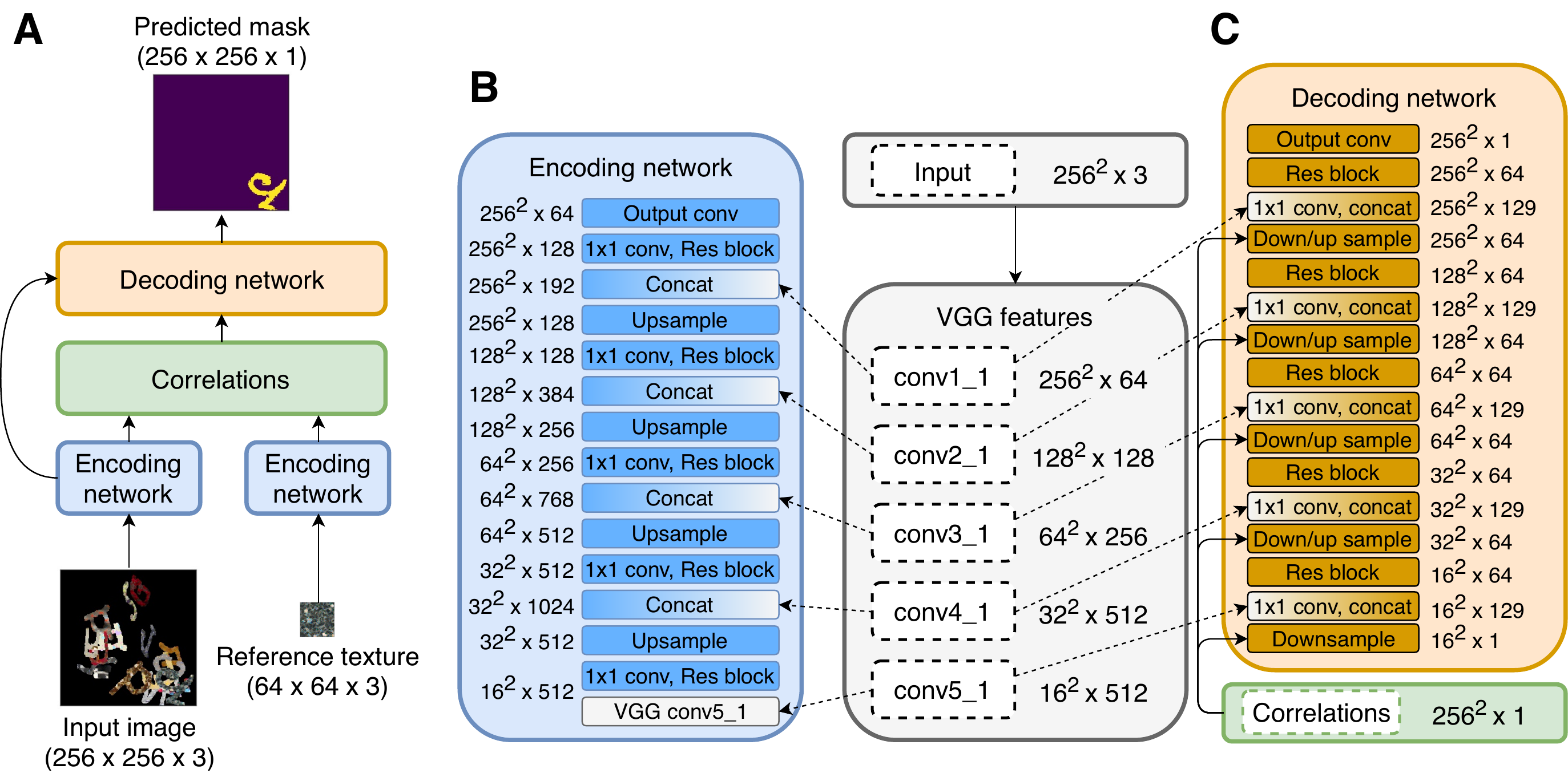}
    \caption{\textbf{A:} Overview of the architecture of the model, \textbf{B:} the architecture of the encoding network, \textbf{C:} the architecture of the decoding network. Residual blocks used in the encoder and decoder are identical (see text for details).}
    \label{fig:architecture}
\end{figure}
\subsection{Texturized cluttered Omniglot}
\label{subsec:cluttered-omniglot}

The second data set we consider is based on texturized Omniglot characters \cite{omniglot2015}. It is generated by uniformly spreading texturized Omniglot characters on a black background. To do so, we take $105 \times 105$ characters from the Omniglot data set, scale them by a factor sampled from $U[0.5, 2]$ and rotate by an angle sampled from $U[0, 2\pi)$. These transformed characters are used as masks for textures, which are then put into a $256 \times 256$ image at random locations with $(x,y)$ coordinates being integers independently sampled from $U[28, 228]$. Examples of the resulting images are in Figure~\ref{fig:results}.

Since Omniglot characters are usually composed of thin lines, we cannot use textures with long-range spatial dependencies (such as natural textures). Therefore we re-synthesize the DTD textures by matching Gram matrices in the first convolutional layer of the VGG network \cite{gatys2015} and use these simpler textures to fill in the Omniglot characters (Figure~\ref{fig:results}). Resulting textures have shorter length scales not exceeding the typical size of an Omniglot character but contain patterns which cannot be modelled by simple colour matching as shown in Sec.~\ref{subsec:non-trivial-embedding}. We reserve 100 randomly chosen textures and characters for testing.

A more challenging version of this data set is obtained by adding random background textures that are different from the textures of the Omniglot characters. The background textures are also re-synthesized DTD textures as described above. An example is given in Figure~\ref{fig:results}.

\section{Model architecture}
\label{sec:proposed-architecture}

We solve the task of one-shot texture segmentation in three steps. First, we compute embeddings of an input image and a reference patch; second, we search for the reference texture in the embedding space to produce a rough segmentation mask; and, finally, we employ a decoding network to produce the output segmentation. The architecture is summarized in Figure~\ref{fig:architecture}.

\subsection{Encoding network}
\label{subsec:encoding-network}

The image embeddings that we are looking for should have two properties: (1) they should map different instances of the same texture to the same point in the embedding space, and (2) they should be local, i.e. for each spatial location encode only the texture in the neighborhood of this location. To ensure (1), we build our encoding network on top of the VGG features \cite{vgg}, which are known to be good texture models \cite{gatys2015}. To ensure (2), the output of the encoding network has the same spatial size as the input, where the feature vector at each spatial position can be thought of as a representation of a local texture at this position.

To compute the embedding of an image (either an input one or a reference patch), we start by extracting the VGG features for this image (specifically, feature maps in layers conv1\_1, conv2\_1, conv3\_1, conv4\_1, conv5\_1). Next, starting with the layer conv5\_1, we repeatedly employ a computational unit consisting of a residual block, upsampling (bilinear interpolation), and concatenation of VGG features of the corresponding resolution to ultimately obtain an embedding of the same spatial resolution as the input. In particular, we use layer conv5\_1 as an input to a residual block, consisting of three convolutional layers with $3 \times 3$ kernels (the first two of which are followed by ReLU non-linearities, while the third one is linear), and then up-sample the output of the residual block by a factor of 2 along each spatial dimension (bilinear interpolation). The resulting tensor has the same spatial dimensions as the VGG layer conv4\_1, which we concatenate to it. Thereafter we repeat the same series of computations (residual block, upsampling, concatenation) until we obtain a tensor of the same spatial resolution as the input, to which we apply a $1 \times 1$ convolutional layer to obtain the final embedding.

The architecture of the encoding network and the exact numbers of feature maps are shown in Figure~\ref{fig:architecture}B.

\subsection{Searching in the encoded space}
\label{subsec:searching-encoded}

Having encoded both the input image and the reference patch, we independently normalize the feature vectors at each spatial position to have norm equal to 1, and spatially convolve the two embeddings with each other. This corresponds to computing cosine distances (up to a constant factor) between them at each spatial position. The resulting feature map of the same spatial size as the input image shows locations with textures similar to the one of the reference patch (see the green boxes in Figure~\ref{fig:architecture}). It is then processed by the decoding network to produce the final segmentation.

\subsection{Decoding network}
\label{subsec:decoding-network}

The decoding network has a similar architecture to the encoding one (Figure~\ref{fig:architecture}C). The main difference is that while the architecture of the residual blocks remains the same, we concatenate not only the VGG features after the upsampling layers, but also the (downsampled) map of cosine distances obtained in the previous step (Section~\ref{subsec:searching-encoded}) Before concatenating the VGG features, we pass them through a $1 \times 1$ convolutional layer to reduce the number of feature maps to 64, which we found to be sufficient. After the final residual layer we apply a $1 \times 1$ convolution with sigmoid non-linearity to produce a single segmentation map ($256 \times 256 \times 1$). The value of each pixel in this map can be interpreted as the probability that this pixel belongs to the reference texture.

\section{Training and Evaluation}
\label{sec:loss-training-evaluation}

\subsection{Loss function}
\label{subsec:loss-function}

We train using standard pixel-wise cross entropy loss \cite{long2015} and apply a weighting term to compensate for the difference in foreground and background size. We denote the ground-truth segmentation mask for the input image $\bx$ as $S(\bx) \in \{0, 1\}^{256 \times 256}$, and the predicted one (i.e. the output of the decoding network, Section~\ref{subsec:decoding-network}) as $\Sh(\bx) \in [0, 1]^{256 \times 256}$. To compare $S(\bx)$ and $\Sh(\bx)$, we use pixel-wise binary cross-entropy loss. Since most of the pixels belong to the background, we separately compute losses for foreground and background and scale them by the corresponding numbers of pixels. Specifically, if the total number of foreground $\Nfg$ and background $\Nbg$ pixels are given by
\begin{equation}
    \Nfg = \sum\limits_{ij}{S_{ij}(\bx)} \quad \text{and} \quad \Nbg = \sum\limits_{ij}{1 - S_{ij}(\bx)},
\end{equation}
we define the loss $\LL(\bx)$ for an image $\bx$ as
\begin{align}
    \LLfg(\bx) & = -\frac{1}{\Nfg} \sum\limits_{ij} S_{ij}(\bx) \log{\hat{S}_{ij}(\bx)}, \\
    \LLbg(\bx) & = - \frac{1}{\Nbg} \sum\limits_{ij} (1 - S_{ij}(\bx)) \log{(1 - \hat{S}_{ij}(\bx))}, \\
    \LL(\bx) & = \LLfg(\bx) + \LLbg(\bx).
\end{align}

\begin{figure}[t]
    \centering
    \includegraphics[width=\textwidth]{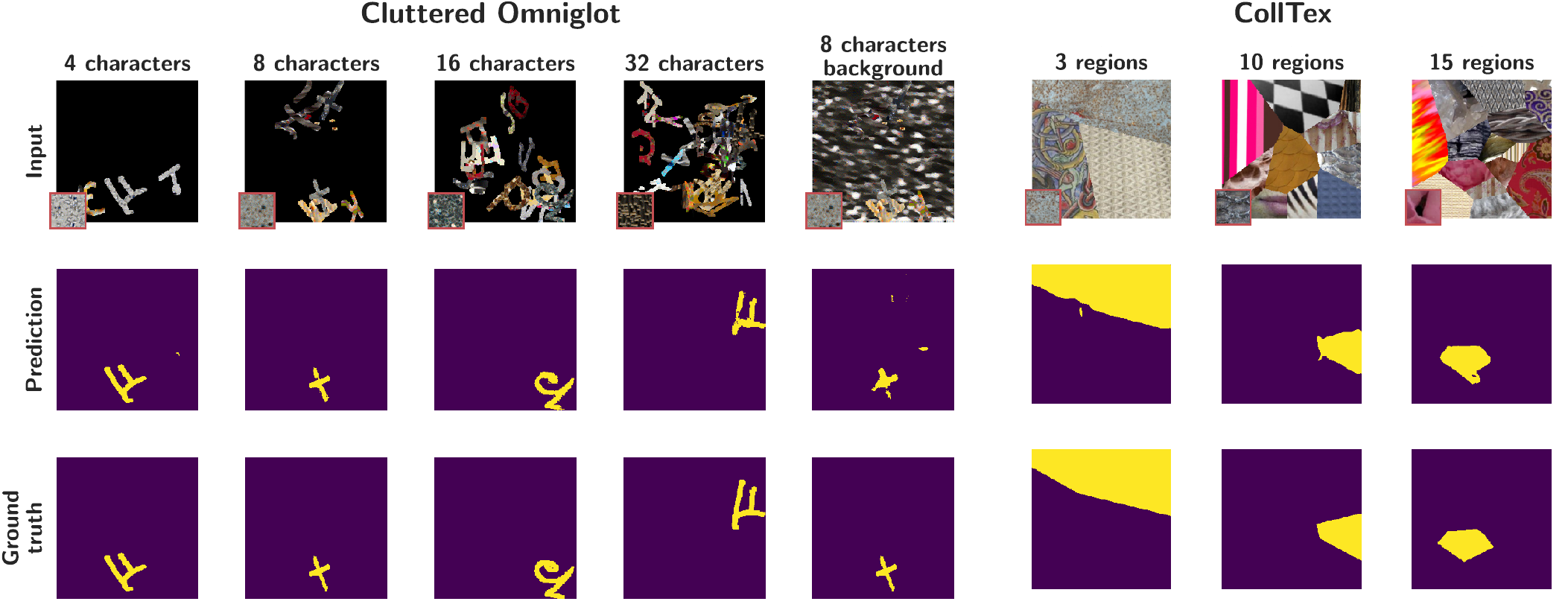}
    \caption{Qualitative results of one-shot texture segmentation for cluttered Omniglot images with different number of characters (left) and CollTex images with different numbers of segmentation regions (right). Input images are shown in the first row, predicted segmentation in the second, and ground truth segmentations in the third. Reference textures are shown in the bottom-left corner of the input image in a red frame.}
    \label{fig:results}
\end{figure}

\subsection{Training}
\label{subsec:training}

The training of the model consists of minimizing the mean loss $\mathbb{E}[\LL(\bx)]$ over the training data set with respect to the parameters of the convolutional layers of the encoding and decoding networks while keeping the weights of the VGG feature extractor network fixed (see Section~\ref{sec:proposed-architecture}). We implement the networks in TensorFlow \cite{tensorflow}, estimate $\mathbb{E}[\LL(\bx)]$ using mini-batches of 8 images, and use Adam optimizer \cite{adam} with all parameters set to default values. We set the learning rate to the initial value of $10^{-5}$ and decrease it by a factor of 2 every $4 \cdot 10^5$ steps.

\subsection{Evaluation}
\label{subsec:evaluation}

We use the Intersection over Union (IoU) as a metric to evaluate the quality of the predicted segmentation masks (binarized using a threshold of $0.9$, which we found to be optimal). All evaluations are done on the held-out textures and characters not used during training.

\section{Results}
\label{sec:results}

In this section we discuss our results on CollTex and texturized cluttered Omniglot, demonstrate the need for a sufficiently capable encoder, and finally, demonstrate two possible applications.

\subsection{Proposed model is a strong baseline}

\begin{figure}[t]
    \centering
    \includegraphics[width=\linewidth]{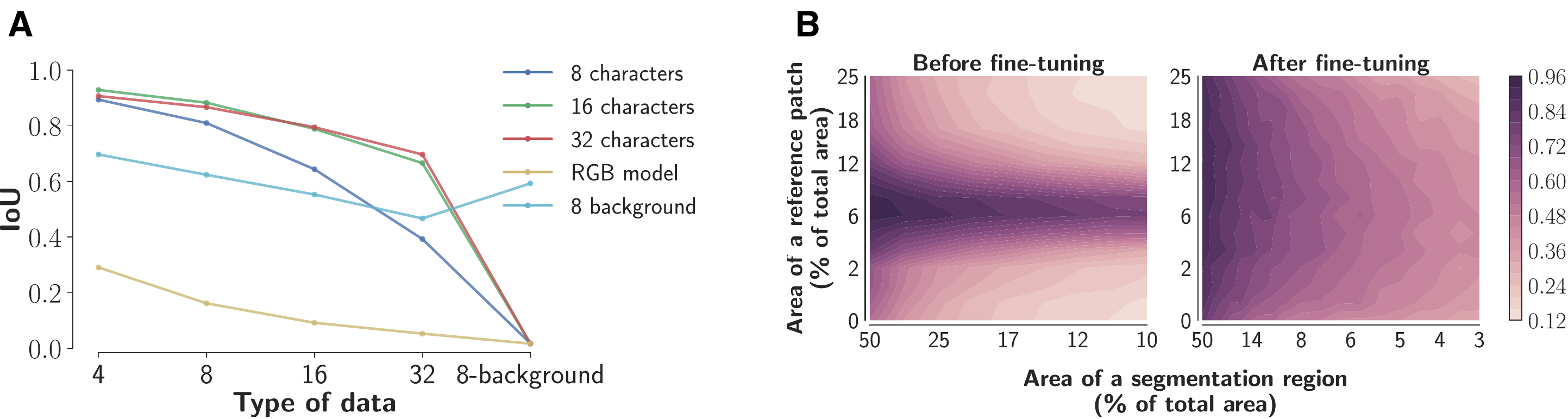}
    \caption{\textbf{A}: IoU scores on the test data for different types of cluttered Omniglot input images for models trained on Omniglot data sets with different numbers of characters. \textbf{B}: IoU scores for the model trained on collages of natural textures before and after fine-tuning on reference patches of variables sizes as a function of a reference patch size and an average area of a segmentation region.}
    \label{fig:iou-plots}
\end{figure}

To train the CollTex model, at each training iteration we draw a random integer $N \in [0, 10]$ and create a random collage of $N$ textures (Section~\ref{subsec:colltex}). After training this model achieves IoU scores between 94\% ($N = 2$) and 76.5\% ($N = 10$) on the held-out set of textures (see more data in Figure~\ref{fig:iou-plots}B(left) and Table~\ref{tab:iou-collages}).

For texturized cluttered Omniglot, we trained separate models on data sets of 8, 16, and 32 characters, as well as on a data set of 8 characters on a background texture. The performance on the images composed of the held-out characters and textures is between 92.9\% (on images with 4 characters) and 69.7\% (images with 32 characters) depending on the number of characters, and 58.9\% for 8 characters with texture background (Figure~\ref{fig:iou-plots}A and Table~\ref{tab:iou-omniglot}).

Similar to what was reported before \cite{claudio2018}, we find that the task becomes harder as we increase the amount of clutter by increasing the number of characters (Figure~\ref{fig:iou-plots}A). The background texture seems to create clutter uniformly throughout the image making the task even more challenging (see IoU scores in Figure~\ref{fig:iou-plots}A and compare the images in the second and fifth columns in Figure~\ref{fig:results}). A similar effect can be seen for collages of natural textures: here the performance decreases as the average area of a segmentation region gets smaller (Figure~\ref{fig:iou-plots}B and and Table~\ref{tab:iou-omniglot}).

\subsection{Encoding network learns non-trivial texture embeddings}
\label{subsec:non-trivial-embedding}

To test whether the networks learn to compute higher order texture statistics, we replaced our encoding network (see Section~\ref{subsec:encoding-network}) with a simpler version consisting of only one linear convolutional layer (with $1 \times 1$ filters), which can only learn to perform a color transformation separately on each pixel. On texturized Omniglot this model (called the ``RGB model'' in Figure~\ref{fig:iou-plots}(a)) performs very poorly compared to the full encoding network. This suggests that the task is not easily solvable in pixel space and the encoding network computes non-trivial and useful texture representations. This result holds for CollTex as well (see Tables~\ref{tab:iou-collages-finetuned} and \ref{tab:iou-collages-rgb}).

\subsection{Pre-training the VGG features extractor is not necessary}
\label{subsec:vgg-pretraining}

The VGG features pre-trained on the ImageNet are known to be good texture models \cite{gatys2015}, therefore we build our model on top of them. However, we would like to test if such pre-training is necessary to solve our task. To do so we randomly initialize the VGG weights, and optimize them jointly with the encoding and decoding networks. We find that a network trained in such a way performs similarly to the one using pre-trained VGG (Tables~\ref{tab:iou-collages} and \ref{tab:iou-collages-no-vgg}), suggesting that the proposed architecture allows to learn texture representations sufficient for the task without any supervision.

\subsection{Training on more difficult data sets helps obtain better models}
\label{subsec:results-best-models}

We evaluated all of our Omniglot models on data sets with 4, 8, 16 and 32 characters. Figure~\ref{fig:iou-plots}(a) shows that the models trained on data sets containing more characters generalize to a smaller number of characters, but not vice versa (best IoU scores are achieved with models trained on 16 and 32 characters). A background texture introduces a different kind of clutter compared to Omniglot characters, making models trained on Omniglot unable to generalize to this case.

\subsection{Model performance is independent of reference patch size}

As described in Section~\ref{sec:one-shot-segmentation-tasks}, we used fixed-size ($64 \times 64$) reference patches for training. That results in size-dependant models with dropping performance as we use reference patches of different sizes (Figure~\ref{fig:iou-plots}(b)). However, fine-tuning the model on reference patches with a randomly chosen size allows to recover the performance for a wide range of patch sizes (Figure~\ref{fig:iou-plots}(b), right panel). We observed the same effect on texturized Omniglot images. This suggests that the proposed model generalizes to different patch sizes and can perform texture matching for patches as small as $16 \times 16$.

\subsection{Demo: Segmenting objects by their textures in videos}
\label{subsec:results-video-segmentation}

\begin{figure}[t]
    \begin{subfigure}{0.48\textwidth}
        \includegraphics[width=\textwidth]{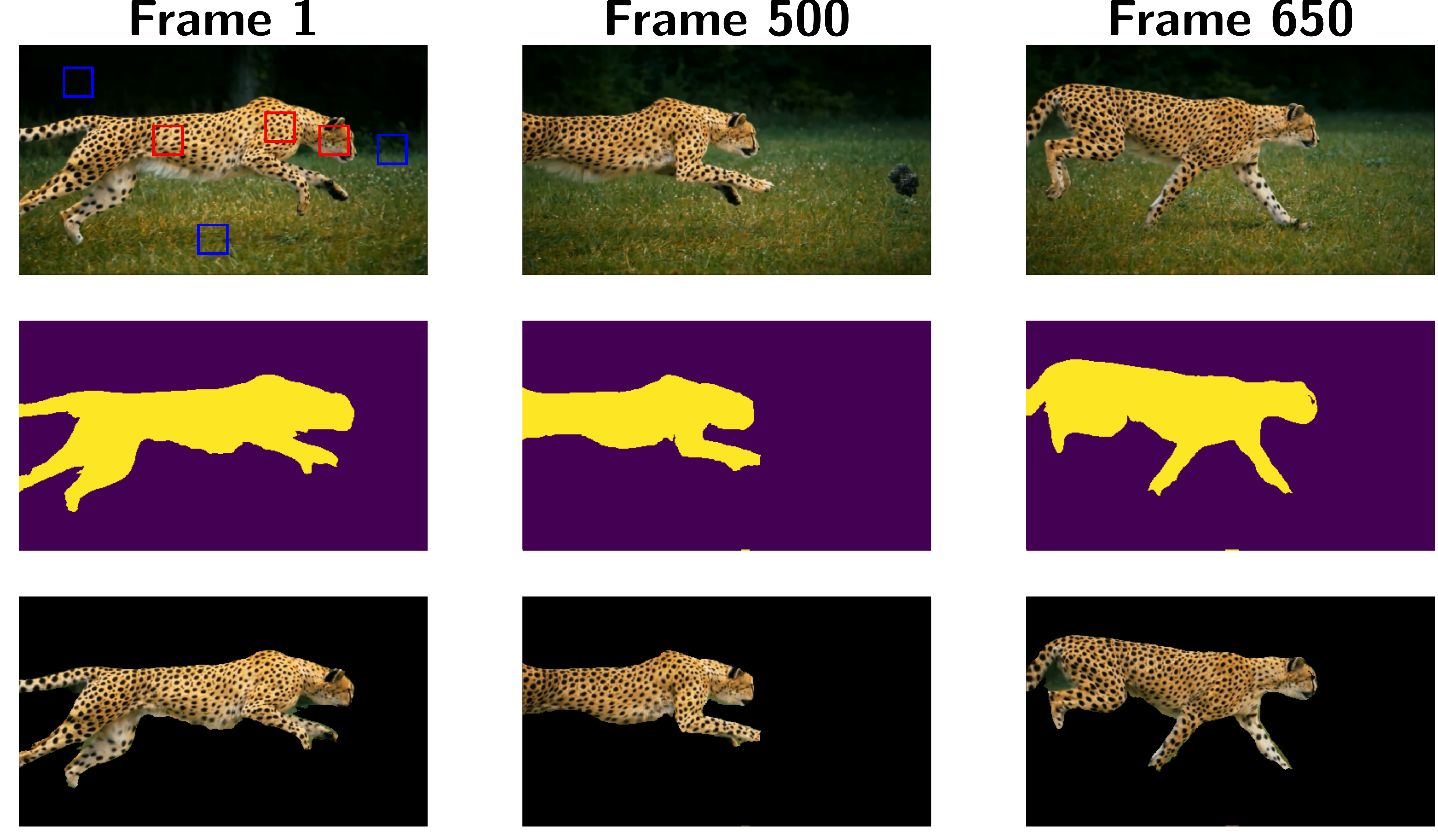}
        \caption{An excerpt from a YouTube video.\footnotemark}
    \end{subfigure}
    \hfill
    \begin{subfigure}{0.48\textwidth}
        \includegraphics[width=\textwidth]{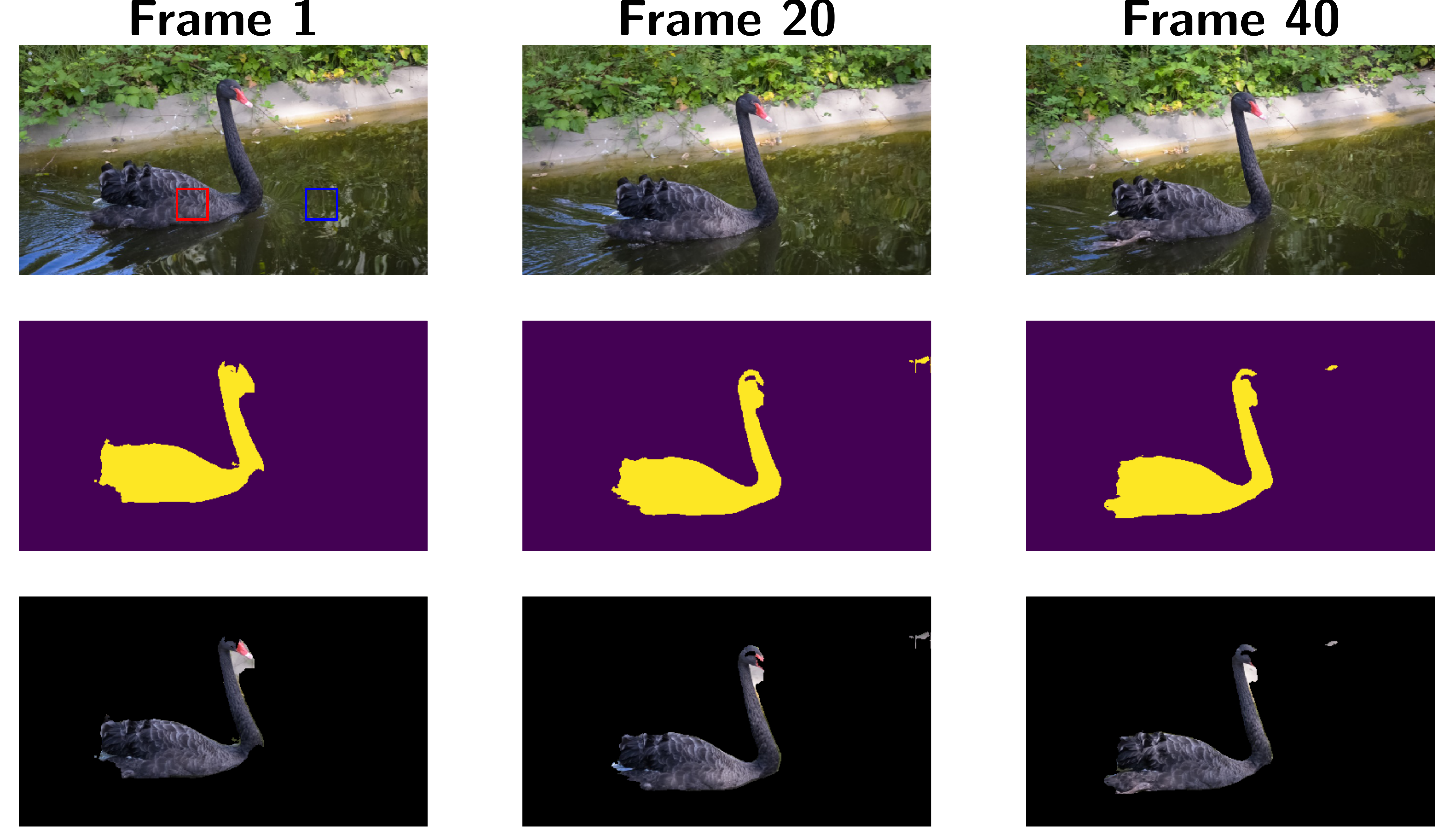}
        \caption{The ``blackswan'' video from DAVIS 2016 \cite{davis2016}.}
    \end{subfigure}
    \caption{Examples of one-shot texture segmentation applied to videos. From the first frame multiple patches containing the texture of an object (shown in red) and background (shown in blue) are extracted. Segmentations for subsequent frames are produced by combining the generated segmentations for each of these patches.}
    \label{fig:video-example}
\end{figure}

\footnotetext{\url{https://youtu.be/B4nd9GF1dRg}}

In Figure~\ref{fig:video-example} we present qualitative results for natural video object segmentation using the model trained on CollTex. From the first frame we extract both the object textures (shown in red) and the background textures (shown in blue). Using them we obtain the segmentation mask of subsequent frames by combining the mask $\Sh_\text{obj}$ for the object texture and the one $\Sh_\text{bg}$ for the background: $\Sh = \Sh_\text{obj} \cdot (1-\Sh_\text{bg})$. The video in Figure~\ref{fig:video-example}B was taken from the DAVIS 2016 data set \cite{davis2016}, which provides the ground-truth segmentations. Using only two texture patches our model reaches an IoU of $84.8\%$ despite being trained exclusively on the synthetic CollTex data set. In comparison, the current state-of-the-art semi-supervised and unsupervised models (which are designed and trained explicitly for this data set) reach the corresponding IoU scores of $96.3$\% and $90.4\%$\footnote[3]{Results from \url{http://davischallenge.org/davis2016/soa_compare.html}}. Further discussion of these results can be found in Section~\ref{sec:discussion}. 

\subsection{Demo: Classifying objects by their textures}
\label{subsec:results-object-classification}

\begin{figure}[t]
    \begin{subfigure}{0.49\textwidth}
        \includegraphics[width=\textwidth]{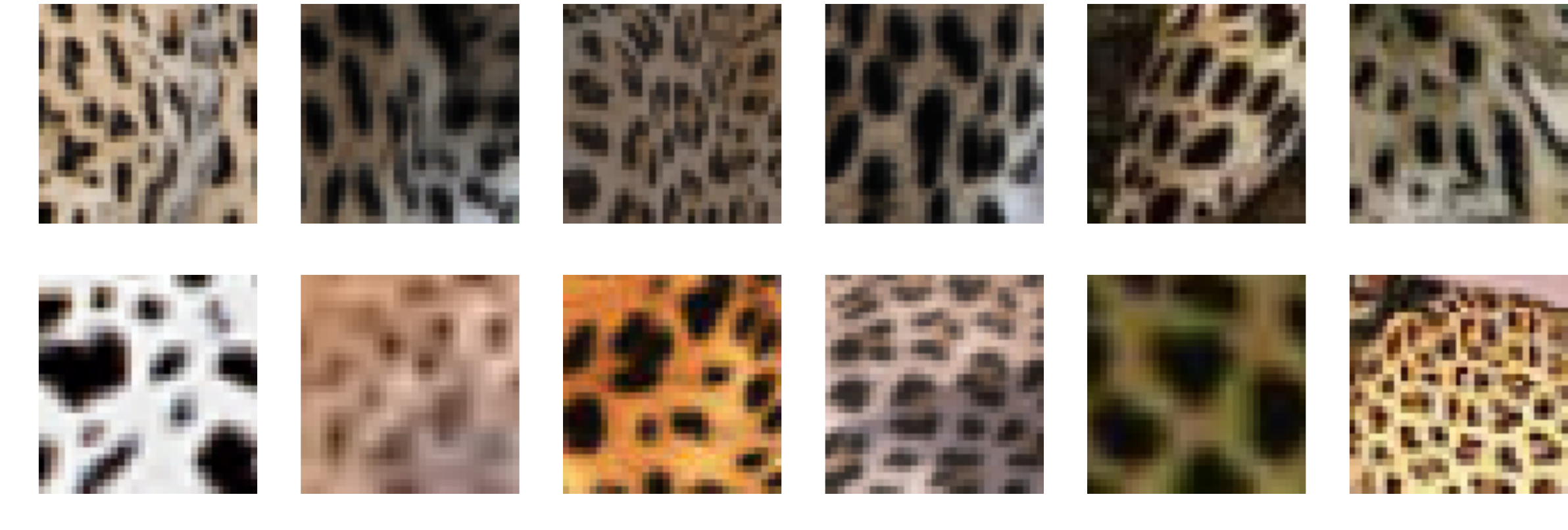}
        \caption{Best (top row) and worst (bottom row) leopard patches for discriminating leopards from other animals.}
    \end{subfigure}
    \hfill
    \begin{subfigure}{0.49\textwidth}
        \includegraphics[width=\textwidth]{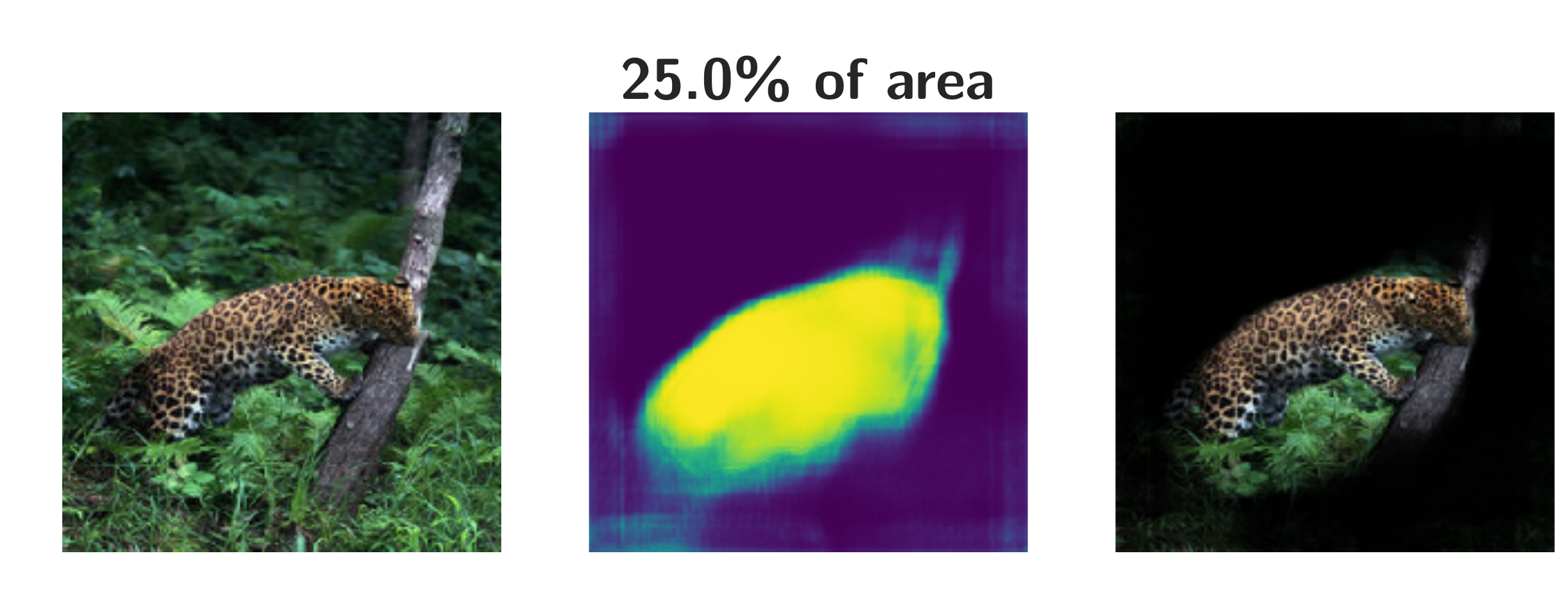}
        \caption{Leopard image segmented using the top-left patch in panel (a) (incorrectly classified: area $\leq$ 30\%).\newline}
    \end{subfigure}
    \begin{subfigure}[b]{0.49\textwidth}
        \includegraphics[width=\textwidth]{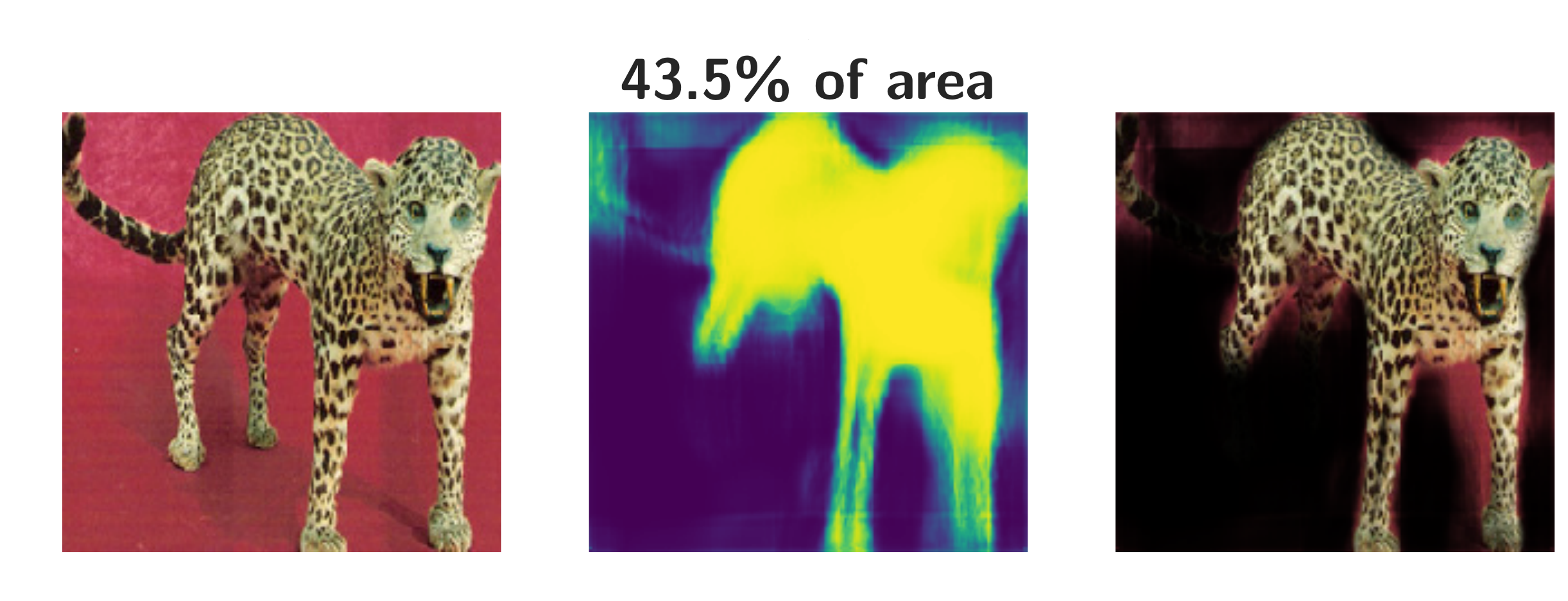}
        \caption{Leopard image segmented using the top-left patch in panel (a) (correctly classified: area > 30\%).}
    \end{subfigure}
    \hfill
    \begin{subfigure}[b]{0.49\textwidth}
        \includegraphics[width=\textwidth]{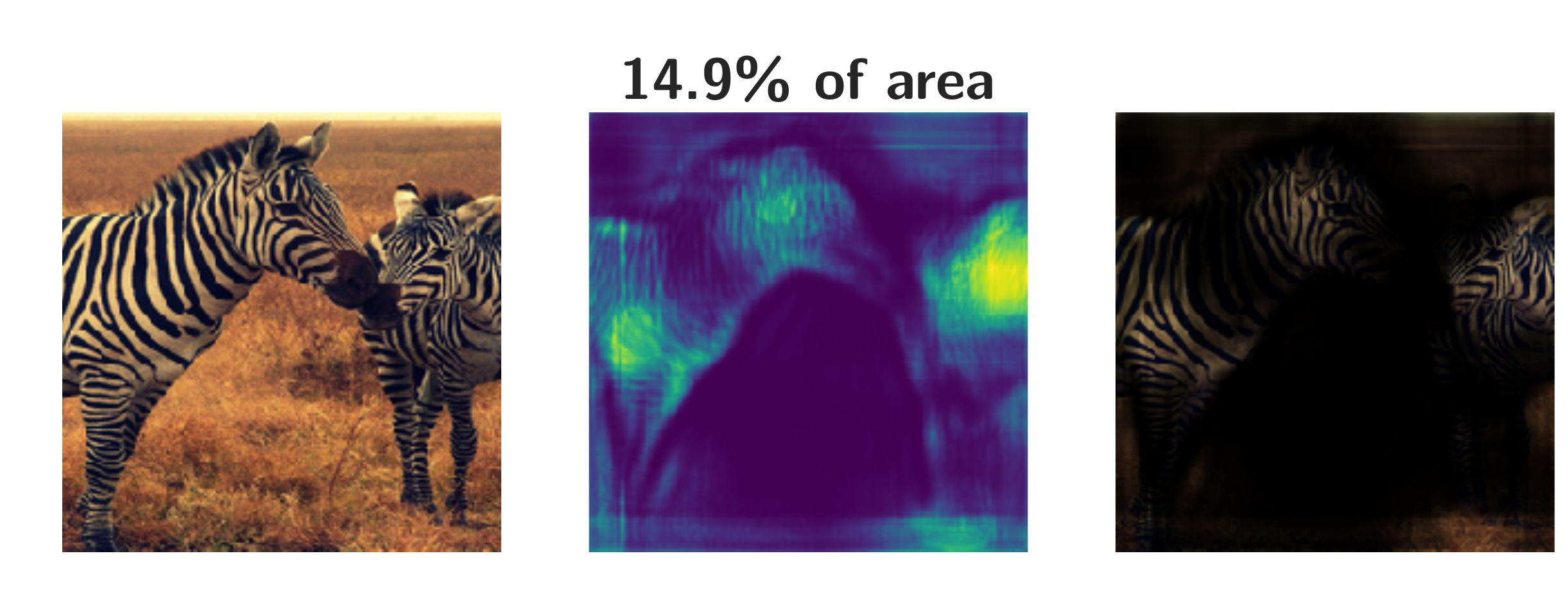}
        \caption{Different class image segmented using the top-left patch in panel (a) (correctly classified: area $\leq$ 30\%).}
    \end{subfigure}
    \hfill
    \caption{Examples of classification of ImageNet images of leopards against images of 4 other ImageNet classes (see details in the text) by their textures.}
    \label{fig:classification-example}
\end{figure}

We extracted images of five different classes (leopard, salamander, zebra, monarch butterfly, hedgehog) from the ImageNet data set \cite{imagenet} to demonstrate that our model trained only on synthetic collages of textures can be used for quick object classification in natural images. For each class we extract the 100 most predictive texture patches (32 $\times$ 32 pixels) using a deep bag-of-features model (manuscript submitted to NIPS, see example patches in Figure~\ref{fig:classification-example}(a)). Using these we perform a one-vs-all classification by selecting those images in which more than $30\%$ of the image is covered by the target texture. For the leopard class this leads to a test performance of $85\%$. In comparison, a linear model on pixel values achieves around $70\%$. In addition to classifying images using the texture patches from the bag-of-features model, we can use the classification accuracy of our model to identify the most and least informative patches (see Figure~\ref{fig:classification-example}(a)). We do so by using each of the extracted patches as a reference patch for the texture segmentation model, and employ the accuracy of a classifier based on these segmentations as a numerical value for how informative the patch is. Upon visual inspection there is little apparent variation among the most informative patches, while the colour noticeably changes across the least informative ones.

\section{Discussion}
\label{sec:discussion}

We introduced the task of one-shot texture segmentation, which we believe is a powerful formalization of texture-based perceptual grouping in terms of a self-supervised learning objective. It allows us to learn useful low- and mid-level representations, which we found to generalize to natural images and videos.

We studied this task on two data sets: (1) a dense collage of multiple textures (CollTex) and (2) texturized cluttered Omniglot. We introduced a strong baseline model to solve these tasks, and demonstrated competitive performance even on task configurations that may be challenging for humans (e.g. 32 characters and 15 regions in Figure~\ref{fig:results}).

As reported in Sections~\ref{subsec:results-video-segmentation} and \ref{subsec:results-object-classification}, our CollTex model generalizes to segmenting objects in natural images and videos, demonstrating its ability to form useful image representations. Even though the performance of our model does not match the current state-of-the-art, it achieves competitive scores without any fine-tuning to the specific task, highlighting its superior generalization properties.

We believe that this work is not only an important addition to self-supervised and one-shot learning, but it also offers new insights into the generalization capabilities provided by the texture representations. We therefore expect the proposed one-shot texture segmentation task to become an important learning target for mid-level representations.

\newpage

\bibliographystyle{plain}
\bibliography{citations}

\appendix

\newpage

\section{Summary of IoU results for cluttered Omniglot and CollTex}

\begin{table}[h]
    \centering
    \begin{tabular}{c r | c c c c c}
         & & \multicolumn{5}{c}{Trained on (number of characters)} \\
         & & 8 & 16 & 32 & 32 RGB & 8-back \\
         \cline{2-7}
         \multirow{6}{*}{Evaluated on} & & & & & & \\
         & 4  & 0.894 & 0.929 & 0.907 & 0.291 & 0.697 \\
         & 8  & 0.810 & 0.883 & 0.867 & 0.162 & 0.624 \\
         & 16 & 0.644 & 0.789 & 0.795 & 0.092 & 0.553 \\
         & 32 & 0.393 & 0.666 & 0.697 & 0.053 & 0.467 \\
         & 8-back & 0.018 & 0.017 & 0.017 & 0.017 & 0.593  
    \end{tabular}
    \caption{IoU scores for the cluttered Omniglot data set with models trained and evaluated on different number of characters (columns correspond to the training data, rows correspond to the evaluation data).}
    \label{tab:iou-omniglot}
\end{table}

\begin{table}[h]
    \centering
    \begin{tabular}{c r | c c c c}
         & & \multicolumn{4}{c}{Type of a model (10 segmentation regions)} \\
         & & Regular & \shortstack{Fine-tuned \\ on variable \\ patch sizes} & \shortstack{No VGG \\ pre-training} & RGB \\
         \cline{2-6}
         \multirow{6}{*}{Reference patch size} & & & & & \\
         & $16 \times 16$   & 0.153 & 0.524 & 0.111 & 0.172 \\
         & $32 \times 32$   & 0.184 & 0.644 & 0.150 & 0.187 \\
         & $64 \times 64$   & 0.765 & 0.692 & 0.693 & 0.169 \\
         & $96 \times 96$   & 0.209 & 0.700 & 0.098 & 0.171 \\
         & $128 \times 128$ & 0.131 & 0.576 & 0.098 & 0.167  
    \end{tabular}
    \caption{IoU scores for models trained on CollTex images with 10 segmentation regions. The columns correspond to different models: regular one, the one fine-tuned on reference patches of variable sizes, the one not using pre-trained VGG, and the RGB model using a pixel-wise color transform for encoding.}
    \label{tab:iou-collages-summary}
\end{table}

\section{Detailed IoU results for CollTex}

\begin{table}[h]
    \centering
    \begin{tabular}{c r | c c c c}
         & & \multicolumn{4}{c}{Number of segmentation regions} \\
         & & 2 & 5 & 10 & 20 \\
         \cline{2-6}
         \multirow{6}{*}{Reference patch size} & & & & & \\
         & $16 \times 16$   & 0.556 & 0.254 & 0.153 & 0.086 \\
         & $32 \times 32$   & 0.661 & 0.322 & 0.184 & 0.113 \\
         & $64 \times 64$   & 0.940 & 0.856 & 0.765 & 0.603 \\
         & $96 \times 96$   & 0.699 & 0.391 & 0.209 & 0.010 \\
         & $128 \times 128$ & 0.590 & 0.256 & 0.131 & 0.062  
    \end{tabular}
    \caption{IoU scores for a model trained on CollTex images (using $64 \times 64$ reference patches).}
    \label{tab:iou-collages}
\end{table}

\begin{table}[h]
    \centering
    \begin{tabular}{c r | c c c c}
         & & \multicolumn{4}{c}{Number of segmentation regions} \\
         & & 2 & 5 & 10 & 20 \\
         \cline{2-6}
         \multirow{6}{*}{Reference patch size} & & & & & \\
         & $16 \times 16$   & 0.829 & 0.658 & 0.524 & 0.432 \\
         & $32 \times 32$   & 0.870 & 0.774 & 0.644 & 0.484 \\
         & $64 \times 64$   & 0.913 & 0.819 & 0.692 & 0.543 \\
         & $96 \times 96$   & 0.917 & 0.845 & 0.700 & 0.509 \\
         & $128 \times 128$ & 0.894 & 0.762 & 0.576 & 0.371  
    \end{tabular}
    \caption{IoU scores for a model trained on CollTex images (using $64 \times 64$ reference patches) and fine-tuned on reference patches of variable sizes.}
    \label{tab:iou-collages-finetuned}
\end{table}

\begin{table}[h!]
    \centering
    \begin{tabular}{c r | c c c c}
         & & \multicolumn{4}{c}{Number of segmentation regions} \\
         & & 2 & 5 & 10 & 20 \\
         \cline{2-6}
         \multirow{6}{*}{Reference patch size} & & & & & \\
         & $16 \times 16$   & 0.528 & 0.215 & 0.111 & 0.067 \\
         & $32 \times 32$   & 0.623 & 0.280 & 0.150 & 0.095 \\
         & $64 \times 64$   & 0.911 & 0.805 & 0.693 & 0.546 \\
         & $96 \times 96$   & 0.498 & 0.200 & 0.098 & 0.050 \\
         & $128 \times 128$ & 0.474 & 0.198 & 0.098 & 0.050  
    \end{tabular}
    \caption{IoU scores for a model trained on CollTex images (using $64 \times 64$ reference patches) without pre-training the VGG features extractor.}
    \label{tab:iou-collages-no-vgg}
\end{table}

\begin{table}[h]
    \centering
    \begin{tabular}{c r | c c c c}
         & & \multicolumn{4}{c}{Number of segmentation regions} \\
         & & 2 & 5 & 10 & 20 \\
         \cline{2-6}
         \multirow{6}{*}{Reference patch size} & & & & & \\
         & $16 \times 16$   & 0.607 & 0.320 & 0.172 & 0.095 \\
         & $32 \times 32$   & 0.627 & 0.325 & 0.187 & 0.101 \\
         & $64 \times 64$   & 0.642 & 0.333 & 0.169 & 0.106 \\
         & $96 \times 96$   & 0.615 & 0.313 & 0.171 & 0.101 \\
         & $128 \times 128$ & 0.620 & 0.306 & 0.167 & 0.089  
    \end{tabular}
    \caption{IoU scores for a model trained on CollTex images (using $64 \times 64$ reference patches) using the encoding network consisting of a single linear $1 \times 1$ convolutional layer.}
    \label{tab:iou-collages-rgb}
\end{table}

\end{document}